\documentclass{IEEEtran}
\usepackage{amsmath}
\usepackage{mathrsfs}
\usepackage{amssymb}
\usepackage{amsthm}

\usepackage{graphicx}
\let\labelindent\relax
\usepackage{enumitem}
\usepackage{epstopdf}
\usepackage{subfigure}
\usepackage{color}
\usepackage{multirow}
 
\usepackage{verbatim}
\usepackage{algorithm,algorithmicx,algpseudocode}
\usepackage{cite}
\usepackage{bbm}
\usepackage{bigstrut}
\usepackage[linktocpage,colorlinks,linkcolor=blue,anchorcolor=blue,citecolor=blue,urlcolor=blue]{hyperref}
\usepackage{cleveref}

\usepackage{tikz}

\makeatletter
\newcommand\footnoteref[1]{\protected@xdef\@thefnmark{\ref{#1}}\@footnotemark}
\makeatother

\usepackage{textcase}
\usepackage[tablename=TABLE]{caption}
\DeclareCaptionTextFormat{up}{\MakeTextUppercase{#1}}
\numberwithin{equation}{section}
\crefname{equation}{equation}{equations}
\crefname{section}{Section}{Sections}

\newtheorem{theorem}{Theorem}
\crefname{theorem}{Theorem}{Theorems}
\crefname{definition}{Definition}{Definitions}

\newtheorem{corollary}{Corollary}
\newtheorem{lemma}{Lemma}
\crefname{lemma}{Lemma}{Lemmas}

\newtheorem{proposition}{Proposition}
\crefname{proposition}{Proposition}{Propositions}

\newtheorem{definition}{Definition}
\crefname{fact}{Fact}{Facts}
\crefname{corollary}{Corollary}{Corollaries}

\crefname{assumption}{Assumption}{Assumptions}
\crefname{algorithm}{Algorithm}{Algorithms}

\crefname{remark}{Remark}{Remark}

\newcommand{\be}{\begin{equation}}
\newcommand{\ee}{\end{equation}}
\newcommand{\ba}{\begin{array}}
\newcommand{\ea}{\end{array}}
\newcommand{\bad}{\begin{aligned}}
\newcommand{\ead}{\end{aligned}}

\usepackage{textcomp}
\usepackage{array,setspace,booktabs,multirow,enumitem}
\usepackage[algo2e]{algorithm2e} 

\def\BibTeX{{\rm B\kern-.05em{\sc i\kern-.025em b}\kern-.08em
		T\kern-.1667em\lower.7ex\hbox{E}\kern-.125emX}}
\IEEEoverridecommandlockouts                        

\title{\LARGE \bf
Structural Estimation of Partially Observable Markov Decision Processes 
}

\author{Yanling Chang, Alfredo Garcia, Zhide Wang, Lu Sun}

\begin{document}
\maketitle

\begin{abstract}

In many practical settings control decisions must be made under partial/imperfect information about the evolution of a relevant state variable.
Partially Observable Markov Decision Processes (POMDPs) is a relatively well-developed framework for modeling and analyzing such problems. In this paper we consider the structural estimation of the primitives of a POMDP model based upon the observable history of the process.
We analyze the structural properties of POMDP model with random rewards and specify conditions under which the model is identifiable without knowledge of the state dynamics. 
We consider a {\em soft} policy gradient algorithm to compute a maximum likelihood estimator and provide a finite-time characterization of convergence to a stationary point. We illustrate the estimation methodology with an application to optimal equipment replacement. In this context, replacement decisions must be made under partial/imperfect information on the true state (i.e. condition of the equipment). We use synthetic and real data to highlight the robustness of the proposed methodology and characterize the potential for misspecification when partial state observability is ignored. 
\end{abstract}
 \IEEEpeerreviewmaketitle
 

\section{Introduction}
\label{intro}
POMDPs generalize Markov Decision Processes (MDPs) by taking into account partial state observability due to measurement noise and/or limited access to information.  When the state is only partially observable (or ``hidden"), optimal control policies may need to be based upon the complete history of implemented actions and recorded observations.
For Bayes optimal control, the updated Bayes belief provides enough information to identify the optimal action. In this context, a POMDP can be transformed to a MDP in which the state is the updated Bayes belief distribution (see \cite{SmallwoodSondik, Monahan, White1991, Krishnamurthy} and many others). POMDP-based models have been successfully applied in a variety of application domains (see e.g. \cite{Winterer, SilverVeness,KrishnamurthyDjonin,FokaTrahanias,YoungMilicaSimon}). 
In this paper we consider the task of estimating the primitives of a POMDP model (i.e. reward function, hidden transition and observation probabilities) based upon the observable histories of implemented actions and observables. 

Under the assumption of complete state observability (for both the controller and the modeler), this problem has been widely studied in two strands of the literature where it is referred to as structural estimation of Markov decision processes (MDP) or alternatively as inverse reinforcement learning (IRL) (see literature review in Section \ref{Lit_Review} below). The partially observable case remains to be considered. The present paper addresses this important gap in the literature. 

In the first part of the paper we introduce a POMDP model with random rewards. For every implemented action, the realization of rewards is observed by the controller but not by the modeler. 
We characterize optimal decisions in a POMDP model with privately observed random rewards
by means of a {\em soft} Bellman equation. This result relies on showing that the updated Bayesian belief on the hidden state is sufficient to define optimal dynamic choices. Here the term {\em soft} is used because optimal policies are characterized by the {\em softmax} function under a specific distributional assumption on random rewards. For a given choice of parameter estimates, the modeler can compute the evolution of latent Bayesian beliefs by the recursive application of Bayes rule for each sample path. Based upon this observation, we introduce a {\em soft} policy gradient algorithm in order to compute the maximum likelihood estimator. Since maximum likelihood is in general a non-convex function, we provide a finite-time characterization of convergence to a stationary point.

We show that the proposed model can be identified if the priori belief distribution, the cardinality of the system state space, the distribution of random i.i.d shocks, the discount factor, and the reward for a fixed reference action are given. We show the estimation is robust to the specification of the a priori belief distribution provided the data sequence is sufficiently long (to address the case where the priori belief distribution is unknown). 

We test and validate the proposed methodology in an optimal engine replacement problem with both synthetic and real data. Using synthetic data, we show that our developed estimation procedure can recover the true model primitives. This experiment also numerically illustrates how model misspecification resulting from ignoring partial state observability can lead to models with poor fit. 
We further apply the model to a widely studied engine replacement real dataset. Compared to the results in \cite{Rust}, our new method can dramatically improve the data fit by $17.7\%$ in terms of the $\log$-likelihood. The model reveals a feature of route assignment behavior in the dataset which was hitherto ignored, i.e. buses with engines {\em believed} to be in worse condition exhibit less utilization (mileage) and higher maintenance costs.

The paper is organized as follows. Section \ref{Lit_Review} provides a literature review with emphasis on how this paper differs from the literature. Section \ref{Sec:Model} introduces a POMDP model with random reward perturbations. Section \ref{Estimation} presents a methodology for structural estimation of POMDP and a policy gradient algorithm. The identification results are presented in Section \ref{Sec:Identification}. Section \ref{sec: numericalexp} provides an illustration to the application of the estimation method to optimal engine replacement problem using both synthetic dataset and the real dataset.  

\section{Literature Review}
\label{Lit_Review}
Research on structural estimation of MDPs features efficient algorithms to address computational challenges in estimating the structural parameters such as determining the value function used in the likelihood function estimation \cite{RustSIAM,Hotz,Hotz1994,Aguirregabiria_1,Aguirregabiria_3,SuJudd,Kasahara}.

In the computer science literature this problem has been studied under the label of inverse reinforcement learning (IRL). A maximum entropy method proposed in \cite{Ziebart} has been highly influential in this literature.
Sample-based algorithms for implementing the maximum entropy method have scaled to scenarios with nonlinear reward functions (see e.g.,\cite{Boularias}, \cite{FinnLevineAbbeel}). In 
\cite{Choi:2011:IRL:1953048.2021028}, the authors extended the maximum entropy estimation method to a partially observable environment, assuming both the transition probabilities and observation probabilities are known with domain knowledge. These methods have also been used for apprenticeship learning where a robot learns from expert-based demonstrations \cite{Ziebart}. To our best knowledge, none of these existing works have developed a general methodology to jointly estimate the reward structure and system dynamics based on observable trajectories of a POMDP process.

\section{A POMDP Model with Random Rewards}
\label{Sec:Model}
At each decision epoch $t\geq 0$, the value of state $s_t \in S$ is not directly observable to the controller nor to the external modeler. However, both the controller and the external modeler are able to observe value of a random variable $z_t \in Z$ correlated with the underlying state $s_t$. We assume the finite action, states and observations. 
If the hidden state is $s_t$ and $a_t \in A$ is implemented, the random reward accrued is $r_{\theta_1}(z_t, s_t,a_t)+\epsilon_t(a_t)$ where $\theta_1 \in \mathbb{R}^{p_1}$ for some $p_1 \in \mathbb{N}_+$ and $\epsilon_t(a_t)$ is a random variable. 
The realization of random reward is observed by the controller but {\em not} by the modeler. This asymmetry of information implies that from the point of view of the modeler, the controller's actions are not necessarily deterministic. 

\begin{figure}[h!]
    \centering
    \includegraphics[width=1.00\linewidth]{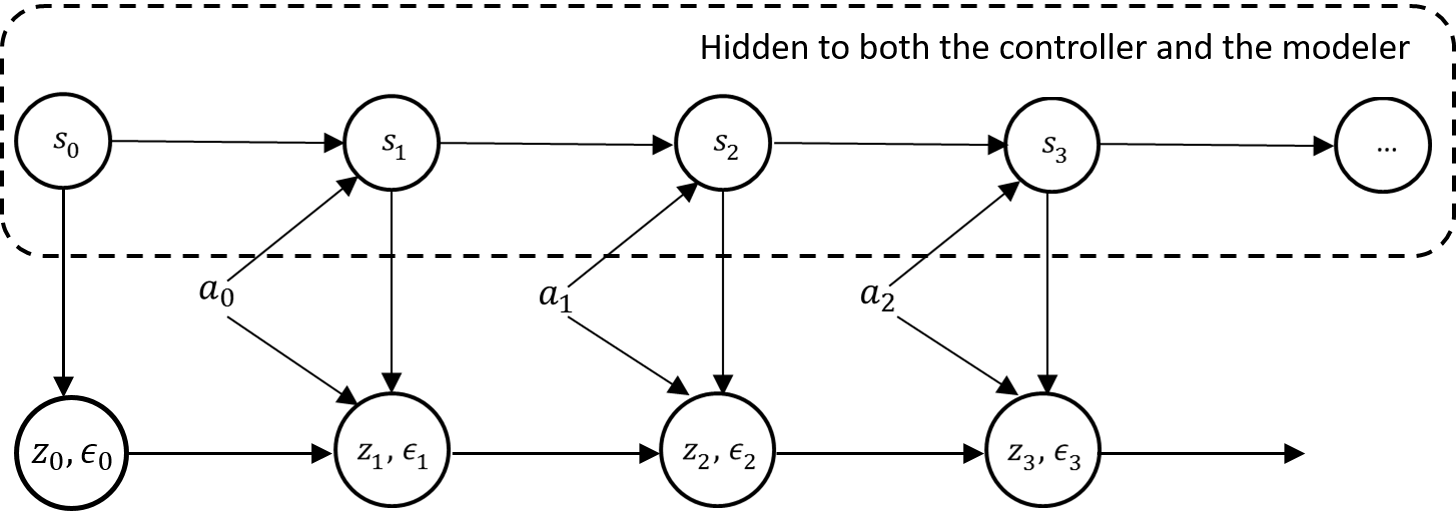}
    \caption{\small{Graphical illustration of the proposed POMDP model with random reward perturbations. At each stage, $z_t$ is observed by both the controller and the modeler, and $\epsilon_t$ is privately observed by the controller. The system state is $s_t$ is hidden to both the controller and the modeler and $s_t$ has its own hidden dynamics.}}
    \label{fig:DMP}
\end{figure}

The system dynamics is described by probabilities $P_{\theta_2}(z_{t+1}, \epsilon_{t+1}, s_{t+1}|{\color{black} z_t},\epsilon_t, s_t, a_t)$ where $\theta_2 \in \mathbb{R}^{p_2}$ for some $p_2 \in \mathbb{N}_+$; see Figure \ref{fig:DMP} for a schematic representation.

Let $\zeta_t = \{z_t,...,z_0, a_{t-1},...,a_0, x_0\}$ be the publicly received history of the dynamic decision process including all past and present revealed observations and all past actions at time $t>0$, where $x_0=\{P(s_0), s_0 \in S\}$ is the prior belief distribution over $S$. The controller aims to maximize 
$$E \left ( \sum_{t=0}^\infty \beta^t [r(z_{t}, s_{t}, a_{t})+\epsilon_t(a_t)]|x_0 \right )$$ 

We assume the following conditional independence (CI): 
\begin{align}
    P_{\theta_2}(z_{t+1},\epsilon_{t+1}|\zeta_t, \epsilon_t, a_t) = P(\epsilon_{t+1}|z_{t+1})P_{\theta_2}(z_{t+1}|\zeta_t, a_t). \label{CIAssumption}
\end{align}
Note that the process $\{\zeta_t, \epsilon_t\}$ is not Markovian; however, $z_{t+1}$ is a sufficient statistic for $\epsilon_{t+1}$, indicating $\epsilon_t$ and $\epsilon_{t+1}$ are independent given $z_{t+1}$ and 
\begin{align}
    P_{\theta_2}(z_{t+1}, &\epsilon_{t+1}, s_{t+1}|{\color{black} z_t},\epsilon_{t}, s_t, a_t)=\notag\\
    & P( \epsilon_{t+1}|z_{t+1})P_{\theta_2}(z_{t+1}, s_{t+1}|z_t, s_t, a_t).
\end{align}
 In addition, the conditional probability $P_{\theta_2}(z_{t+1}|\zeta_t, a_t)$ does not depend on $\epsilon_t$. Intuitively, the CI assumption implies that the POMDP system dynamics $P_{\theta_2}(z_{t+1}, s_{t+1}|z_t, s_t, a_t)$ is superimposed by a noise process $\{\epsilon_t\}$. 
 
 Let $x_{t,\theta_2}=P_{\theta_2}(\cdot|\zeta_t) \in X \subset \mathbb{R}^{|S|-1}$ be the conditional probability distribution given history $\zeta_t$ where $X$ is the unit simplex. Given action $a_t$ and observation $z_t$, the expected reward is: 
 $$
 r_{\theta_1}(z_t,x_{t,\theta_2},a_t)=\sum_{s_t} x_{t,\theta_2}(s_t) r_{\theta_1}(z_t, s_t, a_t).$$
 Under the CI assumption, the optimal decision process can be recursively formulated as:
\begin{align}
  &U_{t,\theta}(\zeta_t, \epsilon_t) = \max_{a_t \in A} \Bigg \{ r_{\theta_1}(z_t,x_{t,\theta_2},a_t)+ \epsilon_t(a_t) \notag \\
  &+\beta \sum_{z_{t+1}}\int P_{\theta_2}(z_{t+1}|\zeta_t, a_t) U_{t+1, \theta}(\zeta_{t+1}, \epsilon_{t+1})d\mu(\epsilon_{t+1}|z_{t+1}) \Bigg \}, \label{model}  
\end{align}
where $\mu(\epsilon_{t+1}|z_{t+1})$ is the cumulative probability distribution of the random perturbation vector $\epsilon_{t+1}$ given the new observation $z_{t+1}$.

The objective for structural estimation is to identify $\theta_1$ in reward $r_{\theta_1}(z_t,s_t,a_t)$, and $\theta_2$ in dynamics $P_{\theta_2}(z_{t+1}, s_{t+1}|z_t, s_t, a_t)$ from the publicly received histories $\{\zeta_T^i\}_{i=1}^N$.

Building upon the POMDP literature \cite{SmallwoodSondik, White1991}, we show in Theorem 1 below that $z_t$ and the updated Bayesian belief distribution $x_{t,\theta_2}$ are sufficient to identify the optimal dynamic choices. 
To this end we mainly follow the notation of  introduce the observation probabilities:
\begin{align*}
&\sigma_{\theta_2}(z_{t+1},{\color{black} z_t}, x_{t,\theta_2}, a_t)  \triangleq
\sum_{s'}\sum_{s}x_{t,\theta_2}(s) P_{\theta_2}(z_{t+1}, s'|{\color{black} z_t},s, a_t), 
\end{align*}
and the belief update function:
\begin{align}
\lambda_{\theta_2}(z_{t+1},{\color{black} z_t},x_{t,\theta_2}, a_t) 
\triangleq \frac{x_{t,\theta_2}P_{\theta_2}(z_{t+1},z_t,a_t)}{\sigma_{\theta_2}(z_{t+1},z_t,x_{t,\theta_2}, a_t)}, \label{lambda} 
\end{align}
assuming $\sigma_{\theta_2}(z_{t+1},z_t,x_{t,\theta_2}, a_t)\neq 0$, where we denote the $(s,s')$ element of the matrix $[P_{\theta_2}(z_{t+1},z_t,a_t)]_{s,s'} \triangleq P_{\theta_2}(z_{t+1}, s'|z_t,s,a_t), s, s' \in S,$ and  \begin{align*}
  &[x_{t,\theta_2}P_{\theta_2}(z_{t+1},z_t,a_t)]_{s_{t+1}} \triangleq 
  \sum_{s}x_{t,\theta_2}(s) P_{\theta_2}(z_{t+1}, s_{t+1}|{\color{black} z_t},s, a_t).  
\end{align*}

\begin{theorem}
Let $\zeta_t$ denote a finite history with current observation $z_t=z$ and updated belief $x_{t,\theta_2}=x$. Let $V_{t,\theta}(z, x, \epsilon)$ be defined as follows:
\begin{align*}
    &V_{t,\theta}(z,x, \epsilon) = \max_{a \in A} \Big\{r_{\theta_1}(z, x, a) + \epsilon(a) \notag \\
    &+ \beta\sum_{z'}\int\sigma_{\theta_2}(z',z,x, a) V_{t+1,\theta}(z',x', \epsilon')d\mu(\epsilon'|z') \Big \}
\end{align*}
where $x' =\lambda_{\theta_2}(z',z,x, a)$. It follows that $V_{t,\theta}(z, x, \epsilon) = U_{t,\theta}(\xi_t,\epsilon)$. Hence, $(z, x)$ is a sufficient statistic for solving \eqref{model}.
\label{SufficientTheorem}
\end{theorem}

\subsection{Soft Bellman Equation}

We now state and prove the {\em soft} Bellman equation for the POMDP model with random reward.
Let $\mathcal{B}$ be the Banach space of bounded, Borel measurable functions $Q: Z\times X \times A \rightarrow R$ under the supremum norm $||.||$. 

Define the {\em soft} Bellman operator $H_{\theta}: \mathcal{B} \rightarrow \mathcal{B}$ by 
\begin{align}
    &[H_{\theta}Q](z,x,a)=r_{\theta}(z,x, a) \notag \\
    &+ \beta \sum_{z'}\sigma_{\theta_2}(z',z,x, a) \int \max_{a \in A} \{ Q(z',x',a)+ \epsilon(a)\}d\mu(\epsilon|z'),
    \label{model4}
\end{align}
where $x' =\lambda_{\theta_2}(z',z,x, a)$.

\begin{theorem}
Under CI and the mild regularity conditions listed in the Appendix, $H_{\theta}: \mathcal{B}\rightarrow \mathcal{B}$ is a contraction mapping with modulus $\beta$. Hence, $H_{\theta}$ has a unique fixed point $Q_{\theta}$ (i.e., $Q_{\theta}=HQ_{\theta}$) and the optimal decision rule $\delta_{\theta}$ is of the form:  
\vspace{-5pt}
\begin{align}
    \delta_{\theta}(z, x,\epsilon) = \arg\max_{a \in A} \Big \{Q_{\theta}(z, x,a) + \epsilon(a)\Big\}.
\end{align}
Furthermore, with conditional choice probabilities 
$$\pi_{\theta}(a|z,x) \triangleq P(a \in \delta_{\theta}(z, x,\epsilon))$$
it holds that $\pi_{\theta}(a|z,x) =\frac{\partial \bar{V}_{\theta}}{\partial Q_{\theta}}$ where:
\begin{align}
 \bar{V}_{\theta}(z,x) & \triangleq \int \max_{a \in A} \{ Q_{\theta}(z,x,a)+ \epsilon(a)\}d\mu(\epsilon|z) \notag \\
 & = \sum_{a \in A} \pi_{\theta}(a|z,x)(Q_{\theta}(z,x,a)+E[\epsilon|a]) \notag \\
 & = E_{a \sim \pi _{\theta}(\cdot |z,x)}
 [Q_{\theta}(z,x,a)+E[\epsilon |a]] .
 \label{Softmax}
\end{align}
\label{InfiniteTheorem}
\vspace{-0.6cm}
\end{theorem}
Finally when the distribution of $\epsilon$ is standard Gumbel, the optimal policy takes a {\em softmax} form.

\begin{theorem}
If the probability measure of $\epsilon$ is multivariate extreme-value, i.e., 
\begin{align}
    \mu(d\epsilon|z)= \prod_{a \in A} \exp\{-\epsilon(a)+\gamma\} \exp\big[-\exp\{-\epsilon(a)+\gamma\}\big],
\end{align}
where $\gamma>0$ is the Euler constant. Then,
\begin{align}
  \pi_{\theta}(a|z,x)=\frac{\exp Q_{\theta}(z,x,a)}{\sum_{a' \in A}\exp Q_{\theta}(z,x,a')},  
\end{align}
where
\begin{align*}
    Q_{\theta}(z,x,a) & =r_{\theta_1}(z,x,a)+\beta\sum_{z'}\sigma_{\theta_2}(z',z,x, a)\bar{V}_{\theta}(z',x'),\\
    x'&=\lambda_{\theta_2}(z',z,x,a),
\end{align*}
and
\begin{align*}
    \bar{V}_{\theta}(z',x') &\triangleq \int V_{\theta}(z',x',\epsilon) d\mu(\epsilon |z') \\
    &= \gamma + \log(\sum_{a'}\exp{Q_{\theta}(z',x', a')}).
\end{align*}
\label{CCPTheorem}
\end{theorem}

\textit{Remark 1}. It can be easily verified that Theorems 1, 2 and 3 continue to hold for the case in which the controller is solving a finite horizon problem. Evidently, the results in this case require that the state-action function $Q_{t, \theta}$ and the conditional choice probabilities $\pi_{t, \theta}$ are time-dependent $t$. 
\section{Maximum Likelihood Estimation} 
\label{Estimation}
Given data corresponding to $N\geq 1$ finite histories of pairs $\{x_{0,i},z_{t,i},a_{t,i}, t=1,...,T\}$ for $i \in \{1,\dots,N\}$, a sequence of trajectories for the belief $\{x_{t,\theta_2,i}:t>0\}$ can be recursively computed for a fixed value of $\theta=(\theta_1,\theta_2)$ as follows:
\begin{align*}
    x_{t+1,\theta_2,i}&=\lambda_{\theta_2} (z_{t+1,i},z_{t,i}, x_{t,\theta_2,i}, a_{t,i}) \\
    &= \frac{x_{t,\theta_2,i}P_{\theta_2}(z_{t+1,i},z_{t,i},
a_{t,i})}{\sigma_{\theta_2}(z_{t+1,i},z_{t,i},x_{t,\theta_2,i},a_{t,i})}.
\end{align*}

Thus, the log-likelihood can be written as:
\begin{align}
   \log \ell(\theta) &\triangleq\log \prod_{i=1}^N P(\zeta_{T,i}|x_{0,i})x_{0,i} \notag\\
   &=\log \prod_{i=1}^N \prod_{t=0}^{T-1}P(z_{t+1,i}|\zeta_{t,i},a_{t,i})P(a_{t,i}|\zeta_{t,i})x_{0,i}\notag\\
    &= \sum_{i=1}^N \sum_{t=0}^{T-1} [ \log \sigma_{\theta_2}(z_{t+1,i},z_{t,i},x_{t,\theta_2,i},a_{t,i}) \notag \\
    &\hspace{10pt}+ \log \pi_{\theta}(a_{t,i}|z_{t,i}, x_{t,\theta_2,i})] + \sum_{i=1}^N\log x_{0,i}. 
    \label{ObjMLE2}
\end{align} 
That is, assuming the data is generated by a Bayesian agent controlling a partially observable Markov process, the external modeler can construct a POMDP model by finding the parameter values that maximize the log-likelihood in \eqref{ObjMLE2}. 

\subsection{A {\em Soft} Policy Gradient Algorithm}

We now introduce a {\em Soft} policy gradient algorithm for approximately maximizing the log-likelihood expression in \eqref{ObjMLE2}. In what follows we shall assume the a priori distribution $x_{0,i}$ is known (see Section \ref{priori} for uncertain/unknown $x_{0,i}$). A two-stage estimator can be obtained by first solving for the
value of $\hat{\theta}_{2}$ that maximizes the value of the first term on
the right hand side in (\ref{ObjMLE2}) and then solve for the value of $\hat{%
\theta}_{1}$ that maximizes the log of pseudo-likelihood $\hat{\ell}(\theta )
$ defined as:%
\[
\hat{\ell}(\theta _{1})=\sum_{i=1}^{N}\sum_{t=0}^{T-1}\log \pi _{(\theta
_{1},\hat{\theta}_{2})}(a_{t,i}|z_{t,i},x_{t,\hat{\theta}_{2},i}).
\]

We simplify notation by using $\pi _{\theta _{1}}$ and $x_{t,i}$ to refer to 
$\pi _{(\theta _{1},\hat{\theta}_{2})}$ and $x_{t,\hat{\theta}_{2},i}$
respectively. For a given value of $\theta _{1}$, consider the {\em soft} Bellman equation:
\begin{align*}
Q_{\theta_1}(z,x,a)& =r_{\theta _{1}}(z,x,a)+\beta 
\sum_{z^{\prime }}\sigma_{\hat{\theta}_{2}}(z^{\prime},z,x,a)\bar{V}_{\theta_1}(z',x'), 
\end{align*}
where $x^{\prime }=\lambda _{\hat{\theta}_{2}}(z^{\prime },z,x,a)$.

After solving the {\em soft} Bellman equation for fixed $\theta_1$ we can compute the
gradient:%
\begin{align*}
&\nabla _{\theta _{1}}\log \pi _{\theta _{1}}(a|z,x) \\
&= \nabla _{\theta _{1}}\log \left (\frac{\exp Q_{\theta_1}(z,x,a)}{\sum_{a' \in A}\exp Q_{\theta_1}(z,x,a')}\right) \\
& =\nabla _{\theta
_{1}}Q_{\theta _{1}}(z,x,a)-\nabla _{\theta _{1}}\log \sum_{a^{\prime }}\exp
Q_{\theta _{1}}(z,x,a^{\prime }) \\
& = \nabla _{\theta
_{1}}Q_{\theta _{1}}(z,x,a)-\nabla _{\theta _{1}} \bar{V}_{\theta_1} (z,x) \\
& = \nabla _{\theta
_{1}}Q_{\theta _{1}}(z,x,a)- \sum_{a'}\pi _{\theta_1}(a' |z,x) \nabla _{\theta _{1}}Q_{\theta_1}(z,x,a').
\end{align*}
The basic steps of a {\em soft} policy gradient algorithm are listed in Algorithm \ref{algo: policy gradient}. Before analyzing the convergence of the {\em soft} policy gradient algorithm we state a preliminary result.

\begin{algorithm}[!ht]
\linespread{1.5}\selectfont
\SetAlgoLined

Compute $\hat{\theta}_{2}$ and $x_{t,i}=x_{t,\hat{\theta}_{2},i}$ $t=1,\ldots T, i=1,\ldots ,N$;
 
Initialize $k=0$, $\theta_{1}^{0}$, $\nabla _{\theta _{1}}\hat{\ell}(\theta _{1}^{0})$, $\epsilon$ and $\rho$;

 \While{\vspace{7pt}$\left\Vert \nabla _{\theta _{1}}\hat{\ell}(\theta_{1}^{k})\right\Vert \ge \epsilon $}{
    
    $k\leftarrow k+1$;
    
    Compute $\nabla _{\theta _{1}} Q_{\theta_1^k} (a|z_{t,i},x_{t,i}), a \in A$;
    
    Compute $\nabla _{\theta _{1}}\hat{\ell}(\theta
    _{1}^{k})=\sum\limits_{i=1}^{N}\sum\limits_{t=0}^{T-1}\nabla _{\theta _{1}}\log \pi_{\theta _{1}^{k}}(a_{t,i}|z_{t,i},x_{t,i})$;
    
    Update parameters $\theta _{1}^{k+1}=\theta _{1}^{k}+\rho \nabla _{\theta _{1}}\hat{\ell}(\theta _{1}^{k})$;
     }
 \caption{\textit{Soft} Policy Gradient Algorithm.}
 \label{algo: policy gradient}
\end{algorithm}

\begin{lemma} \label{bigLemma} Assume $r_{\theta_1}(z,x,a)$ is twice continuously differentiable in $\theta_1 \in \mathbb{R}^{p_1}$ and
\begin{align*}
\sup_{\theta_1}\left\Vert\nabla_{\theta_1}r_{\theta_1}(z,x,a)\right\Vert\leq L_{r,1}<\infty\notag\\ \sup_{\theta_1}\left\Vert\nabla^2_{\theta_1}r_{\theta_1}(z,x,a)\right\Vert\leq L_{r,2} < \infty,
\end{align*}
$\forall (z,x,a) \in  Z \times X \times A$. Then, $Q_{\theta_1}(z,x,a)$ and $\bar{V}_{\theta_1}(z,x)$ are also twice continuously differentiable in $\theta_1 \in \mathbb{R}^{p_1}$ and
\begin{align*}
\sup_{\theta_1}\left\Vert\nabla^2_{\theta_1}Q_{\theta_1}(z,x,a)\right\Vert\leq L_{Q}, & \hspace{0.5cm} \sup_{\theta_1}\left\Vert\nabla^2_{\theta_1}\bar{V}_{\theta_1}(z,x)\right\Vert\leq L_{\bar{V}}    
\end{align*}
$\forall (z,x,a) \in  Z \times X \times A$, where
\begin{align*}
L_Q:=\frac{1}{1-\beta}L_{r,2} 
+\frac{2\beta}{(1-\beta)^3}(L_{r,1})^2 \notag\\
L_{\bar{V}}:= \frac{1}{1-\beta}L_{r,2} 
+\frac{2}{(1-\beta)^3}(L_{r,1})^2.
\end{align*}
\end{lemma}

\begin{theorem}
Under the same assumptions of Lemma 1, the pseudo-log likelihood has Lipschitz continuous gradients with constant $L:=NT(L_Q+L_{\bar{V}})$. With step size $\rho <\frac{2}{L}$ it holds that 
\begin{equation*}
\min_{k\in \{1,\ldots ,K\}}\left\Vert \nabla _{\theta _{1}}\hat{\ell}(\theta
_{1}^{k})\right\Vert ^{2}\leq \frac{1%
}{\rho \Big(1-\frac{\rho L }{2}\Big)}\frac{\hat{\ell}(\theta _{1}^{\ast })-\hat{\ell}
(\theta _{1}^{0})}{K},
\end{equation*}
where $\theta^{*}_1$ maximizes pseudo-log likelihood.
\label{convergenceThm}
\end{theorem}

\section{Model Identification}
\label{Sec:Identification}

The structure of the random reward POMDP model is defined by parameters : 
$$\{r_{\theta_1}(Z, S,A), P_{\theta_2}(Z,S,A), \mu, \beta\},$$
where $r_{\theta_1}(Z,S,A)\equiv\{r_{\theta_1}(z,s,a): z \in Z, s\in S, a \in A\}$, $P_{\theta_2}(Z,S,A)\equiv \{P_{\theta_2}(z',s'|z,s, a), z,z' \in Z, s,s' \in S, a \in A\}$. For the rest of the section, we assume $\mu$ and $\beta$ are given and known. Then the conditional observation probabilities $\{\sigma_{\theta_{2}}(z_{t+1},z_t, x_{t,\theta_2},a_t)\}$ and  conditional choice probabilities $\{\pi_{\theta}(a_t|z_t, x_{t,\theta_2})\}$ are called the reduced form observation probabilities and choice probabilities under structure $\theta=(\theta_1, \theta_2)$.  
Under the true structure $\theta^*=(\theta_1^*, \theta_2^*)$, we must have
\begin{align}
 \forall (z_{t+1}, \zeta_t,a_t), \hspace{40pt}&\notag\\
 \underbrace{ \hat{P}(z_{t+1}|\zeta_t, a_t)}_\textrm{Data} &=\underbrace{\sigma_{\theta_{2}^{*}}(z_{t+1},z_t, x_{t,\theta_{2}^{*}},a_t)}_\textrm{Model},\\
   \underbrace{\hat{P}(a_t|\zeta_t)}_\textrm{Data}  &=\underbrace{\pi_{\theta^*}(a_t|z_t,x_{t,\theta_{2}^{*}})}_\textrm{Model}. 
\end{align}
where $\hat{P}(z_{t+1}|\zeta_t, a_t)$ and 
$\hat{P}(a_t|\zeta_t)$ are functions of the data.

\cite{MagnacThesmar2002} defines the observational equivalence and identification as follows. 
\begin{definition}[Observational equivalence] Let $\Theta$ be the set of structures $\theta$, and let $\stackrel{o}{\Longleftrightarrow}$ be observational equivalence. $\forall \theta, \theta' \in \Theta$, $\theta \stackrel{o}{\Longleftrightarrow} \theta'$ if and only if $$\sigma_{\theta_{2}}(z_{t+1},z_t,x_{t,\theta_2},a_t) = \sigma_{\theta_{2}^{'}}(z_{t+1},z_t,x_{t,\theta'_2},a_t)$$ 
and 
$$\pi_\theta(a_t|z_t, x_{t,\theta_2})=\pi_{\theta'}(a_t|z_t, x_{t,\theta_2'}), \forall z_{t+1},z_t, a_t.$$  
\end{definition}

\begin{definition}[Identification] The model is identified if and only if $\forall \theta, \theta' \in \Theta$, $\theta \stackrel{o}{\Longleftrightarrow} \theta'$ implies $\theta=\theta'$. 

\end{definition}

It is well known that the primitives of a MDP cannot be completely identified in general, and our POMDP model is not an exception. In addition, in the POMDP, the dynamics of the system under study cannot be directly observed as the system state is only partially observable. However, the next theorem shows that we could identify the hidden dynamics using two periods of data (including $x_0$), assuming we know the cardinality of the state space. 
\begin{theorem}
Assume $|S|$ is known. The hidden dynamic $P_{\theta_2}(Z,S, A)$ (not rank-1) can be uniquely identified from the first two periods of data (including $x_0$).
\label{DynamicsIdentification}
\end{theorem}

Theorem \ref{DynamicsIdentification} is crucial as it allows us to generalize the identification results in \cite{Hotz} and \cite{MagnacThesmar2002} for MDPs to POMDPs.

\begin{theorem}
The primitives of the POMDP model cannot be completely identified in general. However, \{$r_{\theta_1}(Z, S,a), a \in A \backslash a^0\}$ and $P_{\theta_2}(Z,S,A)$ can be uniquely identified from the data, given the initial belief $x_0$, the cardinality of the state space $|S|$, the discount factor $\beta$, the distribution of random shock $\mu$, and the rewards $r_{\theta_1}(s, a^0),s \in S$ for a reference action $a^0\in A$ are all known. 
\label{Identification}
\end{theorem}

\textit{Remark 2}. It is well known that the knowledge on the discount factor $\beta$, random shock $\mu$, and the reference reward function is necessary to uniquely identify the primitives of a MDP model \cite{Hotz}. The identification result of the POMDP in Theorem \ref{Identification} only requires two additional mild conditions on the knowledge of the initial belief $x_0$ and the cardinality of the state space, although the system dynamics is hidden and the state is partially observable. We examine the case where the initial belief $x_0$ is unknown in Section \ref{priori}. It is an interesting future research question to examine whether or under what conditions the POMDP model is identifiable if $|S|$ is unknown. In practice, the number of possible states can be obtained by domain knowledge for a particular application. For example, the possible stages of a cancer or system degradation are likely obtainable. A practitioner can also try possible values of $|S|$ to examine which value can best explain the observed behaviors.  
\begin{corollary}
For $T < \infty$, the POMDP model can be uniquely identified from the data, if $x_0, \mu, |S|$ and both the reward structure and the terminal value function $Q_T$ in the reference action $a^0\in A$ are all known. 
\label{corollary1}
\end{corollary}

 \subsection{Sensitivity to A Priori Distribution Specification}
 \label{priori}
 The requirement of knowing the agent's initial belief $x_0$ seems to limit the potential use of the developed estimation approach. However, we now show that the effect of $x_0$ on the estimation result is decreasing with increasing length of the history $\zeta_T$. Specifically, 
 let $\mathcal{D}:X \times X \mapsto \mathbb{R}^+$ be a metric on $X$, defined as
 \vspace{-0.3cm}
\begin{align}
    \mathcal{D}(x,x') &\triangleq \max\{d(x,x'), d(x',x)\},
\end{align}
where 
$$d(x,x')\triangleq 1-\min\left\{\frac{x(s)}{x'(s)}: s \in S, x'(s)>0  \right \}, \forall x,x' \in X.$$ Define 
\begin{align}
  &\eta(P_{\theta_2}(z',z,a)) \notag\\
  &= \max \{\mathcal{D}(\lambda_{\theta_2}(z', z, e_i, a),  \lambda_{\theta_2}(z', z, e_j, a)): i,j \in S\},  
\end{align}
where $e_i \in X$ with 1 on its $i$th element. 
\cite{Platzman} and \cite{White} showed that $\forall x_1, x_2 \in X$, $$\mathcal{D}(\lambda_{\theta_2}(z',z,x_1,a), \lambda_{\theta_2}(z',z,x_2, a)) \leq \eta(P_{\theta_2}(z',z,a))<1,$$ where $\eta(P_{\theta_2}(z',z,a))$ is called a \textit{contraction} coefficient (coefficient of ergodicity) for substochastic matrix $P_{\theta_2}(z',z,a)$. Thus, given a finite history, $\{ z_t, ..., z_{t-M}, a_{t-1}, ..., a_{t-M}\}$, let $\lambda^M$ be $M$ applications of $\lambda$ function for any $x_M \in X$, namely,
\begin{align*}
   \lambda^M_{\theta_2}(z_{t-M}^{t}, &a_{t-M}^{t-1}, x_{t-M}) = \lambda_{\theta_2}(z_{t}, z_{t-1}, \lambda_{\theta_2}(... \\
   &\lambda_{\theta_2}(z_{t-M+1},z_{t-M},x_{t-M}, a_{t-M})), a_{t-1}), 
\end{align*}

where
\begin{align*}
z_{t-M}^{t}=\{z_t,\dots,z_{t-M}\}, a_{t-M}^{t-1}= \{a_{t-1},\dots, a_{t-M}\}
\end{align*} for short. Then
\begin{align*}
    \mathcal{D}(\lambda^M_{\theta_2}(z_{t-M}^{t},&a_{t-M}^{t-1}, x_{t-M}), \lambda^M_{\theta_2}(z_{t-M}^{t},a_{t-M}^{t-1}, x_{t-M}')) \\
    &\leq (\eta(P_{\theta_2}(z',z,a)))^M, \forall x_{t-M}, x_{t-M}' \in X
\end{align*}
(see Section 2.3 in \cite{White}), showing that the effect of $x_0$ decreases as $M$ increases.

 \begin{theorem}
Assume $|S|$ is known. A set of model primitives $\theta$ can be obtained from the data (consistent with an unknown $x_o \in X$), given the discount factor $\beta$, the random shock distribution $\mu$, and the reward $r_{\theta_1}(S,a^0)$. The set of estimators will shrink to the singleton true value as $M \rightarrow \infty$ (hence, $T \rightarrow \infty$). 
\label{x0unknown}
\end{theorem}


Theorem \ref{x0unknown} shows that if $x_0$ is uncertain (to the modeler), we can still estimate a set of $\theta_2$ (hence $\theta_1$) by repeated applications of $\lambda$ function and utilizing the entire data provided by the information sequence $\zeta_T$ (not just the first two-period data), and by varying $x_{t-M} \in X$. The set of estimated $\theta$s will shrink to the singleton true value as $M$ goes to infinity. In addition, in many applications, it is also possible to obtain some (or a small range of) belief points, i.e., $x_0 \in X' \subset X$. For example, in the engine replacement example, it is acceptable to reason that the state of a newly replace engine is good. This information of $X'$ can be very helpful in improving the accuracy of the estimates. 

\section{Illustration: Optimal Equipment Replacement}
\label{sec: numericalexp}

We illustrate the estimation methodology using both synthetic and real data for a bus engine replacement problem. POMDP approaches have been widely used in machine maintenance problems (\cite{Monahan, eckles1968optimum, zixuan2020operational, xu2019optimal}), where maintenance and engine replacement decisions must be made based upon monthly inspection results. Cumulative mileage and other specialized tests only provide informative signals about the true underlying engine's state condition which is not readily observable. 

The engine deterioration state $s_t \in S=\{0, 1\}$, where ``0" is being the ``good state" and ``1" is being the ``bad state". The available actions are $a_t= 1$ is for engine replacement and $a_t= 0$ for regular maintenance.
The model for hidden state dynamics is 
\begin{equation}
P_{\theta_2}(s_{t+1}|s_t,a_t=0) = 
\begin{pmatrix}
\theta_{2,0} & 1-\theta_{2,0} \\
1-\theta_{2,1} & \theta_{2,1}
\end{pmatrix},
\label{dynamic1}
\end{equation}
and $P_{\theta_2}(s_{t+1}=0|s_t,a_t=1)=1$. Per 2500-mile maintenance costs are parametrized by $\theta _{1,0}$ (in good state) and $\theta _{1,1}$ (in bad  state). With a belief $x_t\in (0,1)$ of the engine being in good state and $z_t$ cumulative mileage after $t$ months, the expected (monthly) maintenance cost is
of the form 
\begin{align*}
r_{\theta_1}(z_t,x_t,a=0)&=-0.001[(\theta _{1,0}z_t)x_t+ (\theta _{1,1}z_t)(1-x_t)],\notag 
\end{align*}
and replacement cost is $r_{\theta_1}(z_t,x_t,a=1)=-RC$.
The distribution of monthly mileage increments $\Delta \in \{0,1,2,3\}$ is
parametrized as follows:%
\begin{align}
&P_{\theta _{3}}(\left. z_{t+1}=z_t+\Delta \right\vert
z_t, s_{t}=0,a_{t}=0) \notag \\
&\hspace{38pt}=\theta _{3,0,\Delta }, &&\Delta \in \{0,1,2\}, \notag\\ 
&P_{\theta _{3}}(\left. z_{t+1}=z_{t}+\Delta \right\vert
z_t, s_{t}=0,a_{t}=0)\notag\\
&\hspace{38pt}=1-\theta _{3,0,0}-\theta _{3,0,1}- \theta _{3,0,2}, &&\Delta = 3.%
\label{dynamic2}
\end{align}
Similarly, we define $P_{\theta _{3}}(\left. z_{t+1}=z_{t}+\Delta \right\vert
z_t, s_{t}=1,a_{t}=0)=\theta _{3,1,\Delta }, \Delta \in \{0,1,2,3\}$. Furthermore, after a replacement, the mileage restarts from zero: $P_{\theta _{3}}(\left. z_{t+1}=0 \right\vert
z_t, s_{t},a_{t}=1)=1$. 

\subsection{Synthetic Dataset}
\label{subsec: SyntheticData}
We first show that the developed method can recover the model parameters using synthetic data. In addition, we show how misspecfication errors can arise if an MDP-model (where the state is cumulative mileage) is used to fit data generated by a Bayesian agent in a partially observable environment.

\textbf{Setup.} 
To generate synthetic data, we simulate data with ground truth parameters in Table \ref{table5}. We simulate 3000 buses for 100 decision epochs. Specifically, we randomly generate initial belief $x_0, s_0, z_0$. The selected action is sampled from $\pi_{\theta}(.|x_t,z_t)$ on the basis of current belief $x_t$ and current mileage $z_t$. Once the action is selected, the system state evolves and generates new mileage $z_{t+1}$ according to system dynamics \eqref{dynamic1}-\eqref{dynamic2}. The belief is updated by $\lambda$ function based upon $z_{t+1}, z_t, x_t, a_t$. In this process, only $z_t$ and $a_t$ are recorded and both $x_t$ (except $x_0$) and $s_t$ are discarded (see Algorithm \ref{algo: data generation}).

\vspace{10pt}
\begin{algorithm}[!ht]
\linespread{1.2}\selectfont
\SetAlgoLined
 record = empty holder 
 
 \While{$i < 3000$}{
  randomly generate $x_0, z_0, s_0$;\\
  sample $a_0$ from $\pi_\theta(.|x_0, z_0)$;\\
  record$(i)=[x_0, z_0, a_0]$; 
  
  \While{$t<100$}{
  $z_{t+1}, s_{t+1}$ sampled from $P_{\theta_2, \theta_3}(s, z | s_t, z_t, a_t)$ ;
  
  $x_{t+1} = \lambda(z_{t+1},z_t, x_t, a_t )$;
  
  sample $a_{t+1}$ from $\pi_\theta(.|x_{t+1}, z_{t+1})$;
  
  record=record+$[z_{t+1}, a_{t+1}]$;\tcp{record only $z, a$}
  
  $t \leftarrow t+1$;
  }
  $i \leftarrow i + 1$;
 }
 \caption{Synthetic Data Generation from the POMDP Model.}
 \label{algo: data generation}
\end{algorithm}

\begin{figure*}[!ht]
    \centering
    \includegraphics[scale = 0.3]{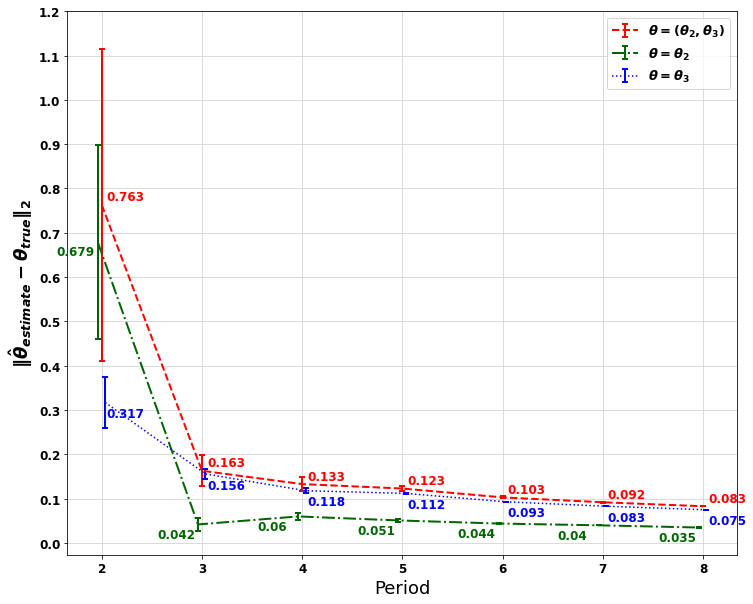}
    \includegraphics[scale = 0.3]{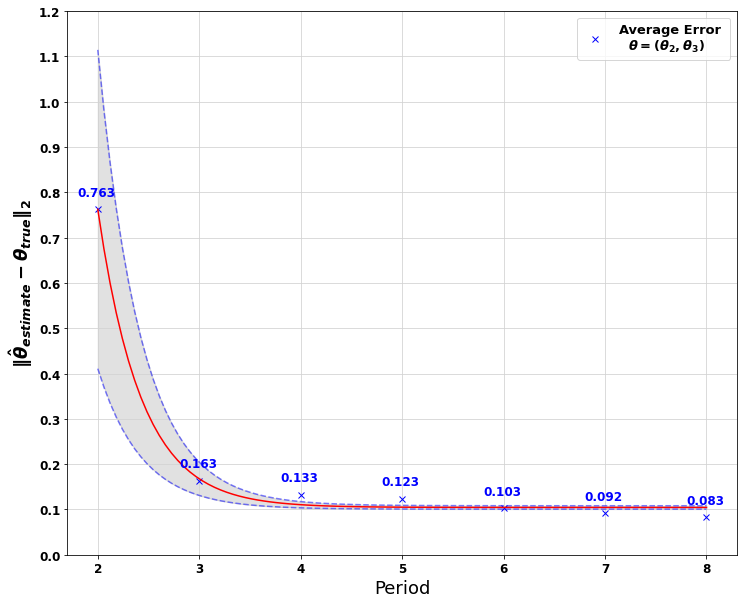}
    \caption{Estimation results are affected by the prior knowledge of $x_0$. Without knowing $x_0$, more periods of data are needed for more accurate estimation. The unknown $x_0$ induced deviation vanishes quickly as the number of data periods grows. }
    \label{fig:x0effect}
\end{figure*}

\textbf{Estimation Results.} The estimation results for the POMDP model are presented in Table \ref{table5}. It shows that our algorithm can identify the model parameters accurately with maximal element-wise deviation of {\color{black}0.006 in dynamics and  0.012 in reward}. In addition, the prior knowledge on initial belief $x_0$ does not influence the estimation result in a significant way (and only improve $\log$-likelihood by {\color{black}0.06\%}). Theorem \ref{x0unknown} shows that if $x_0$ is unknown, the resulting
estimates may deviate from their true values. However, the difference between the estimated results and the true values quickly diminishes as the number of decision epochs increases. Fig. \ref{fig:x0effect} clearly illustrates this fact (where the estimation deviation is caused by varying $x_0 \in X$). For example, with only 8-period of data, the estimated dynamics are within $0.1$ range of the its true value in 2-norm.


\textbf{Model Misspecification}. When the data is generated by a POMDP, using existing MDP-based models will lead to mis-specification errors. To see this, we apply the model in \cite{Rust} to the same synthetic dataset and present the estimation results in Table \ref{table6}. Unsurprisingly, the mis-specification error manifests itself by a significant drop in $\log$-likelihood ($\frac{301750-262973}{301750} = 12.9\%$). In general, the modeling options for a given dataset include MDPs (possibly high-order MDPs), POMDP, or other non-Markovian processes. A central question for the modeler is to select an appropriate model which in the end may not necessarily be Markovian. In this regard, we note that \cite{ShiWan} has recently developed a model selection procedure for testing the Markov assumption. The developed estimation approach for POMDPs can be an appealing alternative when the Markov assumption in sufficiently high order models is still rejected.

\begin{table*}[!ht]
    \centering
        \caption{\small{Parameter estimates and $\log$-likelihood of POMDP-based model.}}
\begin{tabular}{c|ccc ccc cc p{0.7cm} p{0.7cm} c}
    \toprule
    \text{Parameter} & $\theta_{3,0,0}$ & $\theta_{3,0,1}$ & $\theta_{3,0,2}$ & $\theta_{3,1,0}$ & $\theta_{3,1,1}$ & $\theta_{3,1,2}$ & $\theta_{2,0}$ & $\theta _{2,1}$ & $\theta_{1,0}$ & 
    $\theta_{1,1}$ & $RC$ \\ 
    \midrule
\text{True value}& 0.039&0.333 &0.590& 0.181&0.757 &0.061 & 0.949 & 0.988& 0.2& 1.2 & 9.243\\
\midrule
\multirow{2}{*}{\text{$x_0$ known}  }   &0.038&0.327 & 0.596 & 0.181&0.754 &0.064 & 0.950& 0.987 & 0.2 & 1.2 & 9.231\\
\cmidrule{2-12}
& \multicolumn{11}{c}{\text{$\log$-Likelihood:} -262,973}\\ 
\midrule
    
\multirow{2}{*}{\text{$x_0$ unknown}} & 0.038&0.327 &0.596 &0.181& 0.754&0.064&0.950 & 0.987&0.2 & 1.2& 9.230\\  
\cmidrule{2-12}
 & \multicolumn{11}{c}{\text{$\log$-Likelihood:} -262,814} \\ 
\bottomrule
\end{tabular}
    \vspace{10pt}

    \label{table5}
\end{table*}

\begin{table}[!ht]
\centering
\caption{\small{Parameter estimates and $\log$-likelihood provided by the MDP model for the synthetic data.}}
\begin{tabular}{c|>{\centering}p{0.6cm} >{\centering}p{0.6cm} >{\centering}p{0.6cm} >{\centering}p{0.6cm} >{\centering}p{0.4cm} >{\centering\arraybackslash}p{0.6cm}}
    \toprule
    \text{Parameter} & $\theta_{3,0}$ & $\theta_{3,1}$ & $\theta_{3,2}$ & $\theta_{3, 3}$& $\theta_{1}$  & $RC$ \\ 
    \midrule
    \multirow{2}{*}{MDP Model} & 0.128& 0.601& 0.257& 0.015& 1.1 & 9.811 \\ 
    \cmidrule{2-7}
     & \multicolumn{5}{c}{\text{$\log$-Likelihood: } -301,750} & \\ 
     \bottomrule
\end{tabular}
\label{table6}
\end{table}

\subsection{Real Dataset}
\label{subsec: RustData}

To illustrate the application of the developed methodology, we now revisit a subset of dataset reported in \cite{Rust}. Specifically, Group $4$ consisting of buses with 1975 GMC engines. Evidence of positive serial correlation in mileage increments is quite strong as the Durbin-Watson statistic is less than $1.13$ for all buses except one with a value of $1.32$. Thus, we fit the data by the POMDP-based model as in Section \ref{subsec: SyntheticData}; see Table \ref{table1}.

\textbf{Discussion}. 
We compared our estimation results (from the POMDP model) with the results found in \cite{Rust} (from a MDP-based model). In terms of log-likelihood, our POMDP-based model outperformed the MDP-based model by $\frac{4495 - 3819}{3819} = 17.7\%$ (see Table \ref{table1} and Table \ref{table2}). 

\begin{table*}[!ht]
\centering
\caption{\small{Parameter estimates and $\log$-likelihood provided by the POMDP-model (standard errors obtained by bootstrapping method are in parentheses)}}

\begin{tabular}{c|ccc ccc cc >{\centering}p{0.7cm} >{\centering}p{0.7cm} c} 
    \toprule
    \text{Parameter} & $\theta_{3,0,0}$ & $\theta_{3,0,1}$ & $\theta_{3,0,2}$ & $\theta_{3,1,0}$ & $\theta_{3,1,1}$ & $\theta_{3,1,2}$ & $\theta_{2,0}$ & $\theta _{2,1}$ & $\theta_{1,0}$ & 
    $\theta_{1,1}$ & $RC$ \\ 
    \midrule
    
  {\text{Good State}} & 0.039& 0.335 & 0.588 & {$\ast$ }&{$\ast$} &
    {$\ast$} & 0.949 & {$\ast$} & 0.3 & {$\ast$} & \\
    
    &(.005)& (.018) & (.018) &&& & (.004) &  & (.3) &  &{9.738}  \\
    
{\text{Bad State}} & {$\ast$ }&{$\ast$} &
   {$\ast$} &0.182& 0.757&0.061 & {$\ast$}&0.988 & {$\ast$} & 1.3& {(1.052) } \\
    
     & && & (.008) & (.008) & (.006) &  & (.002) & & (.2) &  \\ 
     \midrule
    \text{$\log$-Likelihood} & \multicolumn{10}{c}{ -3819} & \\ 
    \bottomrule
\end{tabular}

\vspace{10pt}

\label{table1}
\end{table*}  

\begin{table*}[!ht]
\centering
\caption{\small{Parameter estimates and $\log$-likelihood with MDP Model}}
\begin{tabular}{c|c c c c >{\centering}p{1cm} c}
    \toprule
    \text{Parameter} & $\theta_{3,0}$ & $\theta_{3,1}$ & $\theta_{3,2}$ &$\theta_{3,3}$& $\theta_{1}$  & $RC$ \\ 
    \midrule
    \text{MDP Model \cite{Rust} p. 1022} & 0.119 & 0.576 & 0.287 & 0.016& 1.2 & 10.90 \\ 
 (Standard errors in parentheses)   & (0.005) & (0.008) & (0.007) & (0.002) &(0.3) & (1.581) \\
    \midrule
    \text{$\log$-Likelihood} & \multicolumn{5}{c}{-4495} & \\
    \bottomrule
\end{tabular}
\vspace{10pt}

\label{table2}
\end{table*}

Compared to the MDP model displayed in Table \ref{table2}, the POMDP model also captures a feature of engine utilization: the distribution of mileage increments for engines considered in bad state is dominated (in the first-order stochastic sense) by the distribution of mileage increments of engines in good state. Furthermore, we find that the marginal operation costs ($\theta_{1, 1}$) for buses in {\em bad} state is significantly higher than those ($\theta_{1, 0}$) in {\em good} state (at least about two times, taking the standard deviation into consideration). 

To gauge the economic interpretation of this result, we follow the same scaling procedure as in \cite{Rust} to get the dollar estimates for $\theta_{1,0}$ and $\theta_{1,1}$, respectively. All estimates are scaled with respect to reported (average) replacement cost (in 1985 US dollars) which for group 4 is $\$7513$ (see Table III in \cite{Rust}, p. 1005).
The {\em perceived} average monthly maintenance costs increases $\frac{7513}{RC}\theta_{1,0}=\$0.231$ per 2500 miles in good state and $\frac{7513}{RC}\theta_{1,1}=\$1$ in bad state. That is, an engine with $300K$ miles in {\em good} condition has (according to the model) a monthly maintenance cost of $(300/2.5)\times 0.231=\$27.6$ whereas an engine with $300K$ miles in {\em bad} condition has monthly maintenance cost of $(300/2.5)\times 1.00=\$120$.




\section{Conclusions}
\label{Sec:Conclusions}

In this paper, we developed a novel estimation method to recover the primitives of a POMDP model based on observable trajectories of the process. 
First, we provide a characterization of optimal decisions in a POMDP  model with random rewards by mean of {\em soft} Bellman  equation. We then developed a soft policy gradient algorithm to obtain the maximum likelihood estimator. We also show that the proposed model can be identified if the a priori belief distribution, the cardinality of the system state space, the distribution of random i.i.d shocks, the discount factor, and the reward for a fixed reference action are given. Moreover, we show the estimation is robust to the specification of the a priori belief distribution provided the data sequence is sufficiently long (to address the case where the priori belief distribution is unknown). 
Finally we provide a numerical illustration with an application to optimal equipment replacement. With synthetic data, we show that highly accurate estimation of the true model primitives can be obtained despite having no prior knowledge of the underlying system dynamics. We also compared our POMDP approach to an MDP approach using a real data on the bus engine replacement problem. Our POMDP approach significantly improved the $\log$-likelihood function and also revealed economically meaningful features that are conflated in the MDP model.

As this research represents a first effort on developing estimation methods for partially observable systems, future research directions are numerous. For example, computational challenges of our model are obviously not trivial. POMDPs suffer from the well-known curse of dimensionality, and observations in many real applications can be high dimensional. Thus, a research direction is to address computational challenges of high dimensional hidden state models. This could be done for example via projection methods \cite{zhou} or variational inference \cite{Igl}.

\section{APPENDIX}
\subsection{{\em Soft} Bellman Equation for POMDPs}
\proof{Proof of Theorem \ref{SufficientTheorem}.}
The proof is by induction.

Assume $U_{t+1, \theta}(\zeta_{t+1}, \epsilon_{t+1}) = V_{t+1, \theta}(z_{t+1}, x_{t+1}, \epsilon_{t+1})$, then 

\begin{align*}
  &U_{t, \theta}(\zeta_t, \epsilon_t)\\
  &= \max_{a_t \in A} \Bigg\{ \sum_{s_t}P_{\theta_2}(s_t|\zeta_t) r_{\theta_1}(z_t, s_t, a_t) + \epsilon_t(a_t) + \beta\sum_{z_{t+1}}\int ...\\
  \end{align*}
  \begin{align*}
  &\hspace{20pt}P_{\theta_2}(z_{t+1}|\zeta_t, a_t) V_{t+1, \theta}(z_{t+1}, x_{t+1}, \epsilon_{t+1})d\mu(\epsilon_{t+1}|z_{t+1}) \Bigg \} \\
  &= \max_{a_t \in A} \Bigg \{  r_{\theta_1}(z_t, x_t, a_t) + \epsilon_t(a_t) + \beta \sum_{z_{t+1}}\int ... \\ &\hspace{20pt}\sigma_{\theta_2}(z_{t+1},z_t, x_t, a_t) V_{t+1, \theta}(z_{t+1},\lambda(z_{t+1},z_t, x_t, a_t), \epsilon_{t+1})...\\
  &\hspace{20pt}d\mu(\epsilon_{t+1}|z_{t+1}) \Bigg \} \\
  &=V_{t,\theta}(z_t, x_t, \epsilon_t)
\end{align*}
where the second equality follows from 
$\sigma_{\theta_2}(z_{t+1},z_t,x_t, a_t) = P_{\theta_2}(z_{t+1}|\zeta_t, a_t)$. 
\endproof

\proof{Proof of Theorem \ref{InfiniteTheorem}.}
 Assume the following regularity conditions:
\begin{enumerate}
\item[{\bf R.1}] (Bounded Upper Semicontinuous) For each $a \in A$, $r_{\theta}(z,x, a)$ is upper semicontinuous in $z$ and $x$ with bounded expectation and
   \begin{align*}
        h_{\theta}(z,x) &\triangleq \sum_{t=1}^{\infty}\beta^t h_{t,\theta}(z,x) < \infty, \\
        h_{1,\theta}(z,x) &= \max_{a \in A} \sum_{z'\in Z} \sigma_{\theta_2}(z',z, x,a) \int \max_{a' \in A} \{r_{\theta}(z',x', a')\\
        &\hspace{10pt}+ \epsilon'(a')\} d \mu(\epsilon'|z'),\\
        h_{t,\theta}(z,x) &= \max_{a \in A} \sum_{z' \in Z} \sigma_{\theta_2}(z',z, x, a) h_{t-1}(z',x'); 
    \end{align*}
 where $x' = \lambda_{\theta_2}(z',z, x, a)$.
    
   \item[{\bf R.2}] (Weakly Continuous) The stochastic kernel $$\sigma_{\theta_2}(\cdot,z, x,a)=\{\sigma_{\theta_2}(z',z, x, a)\}_{z' \in |Z|}$$ is weakly continuous in $Z\times X \times A$;
    
    \item[{\bf R.3}] (Bounded Expectation) The reward function $r_\theta \in \mathcal{B}$ and for each $Q \in \mathcal{B}$, $E_{\theta}Q \in \mathcal{B}$, where
    \begin{align*}
         [E_{\theta}Q](z,x, a) =& \sum_{z' \in Z} \sigma_{\theta_2}(z',z, x, a)\\
         &\times\int \max_{a \in A}\{ Q(z',x',a)+ \epsilon(a)\}d\mu(\epsilon|z')
   \end{align*}
 where $x' = \lambda_{\theta_2}(z',z, x, a)$.
\end{enumerate}
Under these regularity conditions, $H_\theta: \mathcal{B} \rightarrow \mathcal{B}$ is well defined (see related discussion in \cite{RustSIAM}). $\forall Q, Q' \in \mathcal{B}$, $\forall a$, we have 
\begin{align*}
    &\Vert H_\theta Q - H_\theta Q' \Vert \\
    &\leq \beta \sum_{z'}\sigma_{\theta_2}(z',z,x, a) \int |\max_{a \in A} \{ Q(z',x',a)+ \epsilon(a)\} \\
    &\hspace{10pt}-\max_{a \in A} \{ Q'(z',x',a)+ \epsilon(a)\}|d\mu(\epsilon|z')\\
   & \leq \beta \sum_{z'}\sigma_{\theta_2}(z',z,x, a) \int \max_{a \in A} |\{ Q(z',x',a)\\
   &\hspace{10pt}-Q'(z',x',a)\}|d\mu(\epsilon|z')\\
   &\leq \beta \Vert Q-Q' \Vert. 
\end{align*}
Hence, $H_\theta$ is a contraction mapping and $Q_\theta$ is the fixed point of $H_\theta$. 
Note that the controlled process $\{z_{t+1}, z_t, x_t, a_t\}$ is Markovian because the conditional probability of $a_{t}$ is $P(a_{t}|z_t, x_{t})$, the conditional probability of $x_{t+1}$ is provided by $\lambda(z_{t+1}, z_t, x_t, a_t)$, and  the conditional probability of $z_{t+1}$ is given by $\sigma(z_{t+1},z_t, x_t, a_t)$. In addition, 
$$\Bigg \Vert \frac{\partial (\max_{a \in A}[ Q_{\theta}(z,x,a)+ \epsilon(a)])}{\partial Q_{\theta}(z,x,a)}\Bigg \Vert\leq 1$$ 
for almost all $\epsilon$, by the Lebesgue dominated convergence theorem, 
\begin{align*}
   \frac{\partial \bar{V}_{\theta}}{\partial Q_{\theta}} &=\int\Bigg (\frac{\partial (\max_{a \in A}[ Q_{\theta}(z,x,a)+ \epsilon(a)])}{\partial Q_{\theta}(z,x,a)} \Bigg ) d\mu(\epsilon|z)\\
      &=\int I\{ a= \arg\max_{a \in A}[ Q_{\theta}(z,x,a)+ \epsilon(a)]\} d\mu(\epsilon|z)\\
      &= \pi_\theta(a|z,x).
\end{align*}
\endproof

\proof{Proof of Theorem \ref{CCPTheorem}.}
The result follows by Theorem \ref{InfiniteTheorem} and \cite{Train} that $$E_\epsilon[\max_a Q_{\theta}(z,x,a) + \epsilon_a] = \gamma + \ln\Big( \sum_{a \in A} \exp{\big( Q_{\theta}(z,x,a) \big)} \Big).$$
\endproof

\subsection{Convergence of {\em Soft} Policy Gradient Algorithm}

\proof{Proof of Lemma \ref{bigLemma}.}~{}

For fixed $\theta_1 \in \mathbb{R}^{p_1}$, consider the mapping $G^1_{\theta_1}: \mathcal{B} \mapsto \mathcal{B}$ defined as:
\begin{align*}
    [G^1_{\theta_1}g]&(z,x,a) = \nabla_{\theta_1}r_{\theta_1}(z,x,a)\\
    & + \beta\sum\limits_{z'}\sigma_{\hat{\theta}_2}(z',z,x,a)\sum\limits_{a'}\pi_{\theta_1}(a'|z',x')g(z',x',a'),
\end{align*}
$$$$ 
where $x' = \lambda_{\hat{\theta}_2}(z',z,x,a)$. It follows that $G^1_{\theta_1}$ is a contraction map with unique fixed point $\nabla_{\theta_1}Q_{\theta_1}$ and $\left\Vert\nabla_{\theta_1}Q_{\theta_1}(z,x,a)\right\Vert
\leq
\frac{1}{1-\beta}L_{r,1}$. Since
\begin{align*}
     \nabla_{\theta_1}\bar{V}_{\theta_1}(z',x') 
&=\sum\limits_{a'}\pi_{\theta_1}(a'|z',x')\nabla_{\theta_1}Q_{\theta_1}(z',x',a')
\end{align*}
it follows that
$\nabla_{\theta_1}\bar{V}_{\theta_1}$ also exists and $\left\Vert\nabla_{\theta_1}\bar{V}_{\theta_1}\right\Vert
\leq
\frac{1}{1-\beta}L_{r,1}$.

Consider the mapping $G^2_{\theta_1}:\mathcal{B} \mapsto \mathcal{B}$ defined as follows:
\begin{align*}
        [G^2_{\theta_1}g]&(z,x,a) = \nabla^2_{\theta_1}r_{\theta_1}(z,x,a)+ \beta\sum\limits_{z'}\sigma_{\hat{\theta}_2}(z',z,x,a) \\
        &\times \sum\limits_{a'}\nabla_{\theta_1}\pi_{\theta_1}(a'|z',x')\nabla_{\theta_1}Q_{\theta_1}(z',x',a') \\
     &+ \beta\sum\limits_{z'}\sigma_{\hat{\theta}_2}(z',z,x,a)\sum\limits_{a'}\pi_{\theta_1}(a'|z',x')g(z',x',a'),
\end{align*}
where $x' = \lambda_{\hat{\theta}_2}(z',z,x,a)$.
Thus, $G^2_{\theta_1}$ is a contraction map with unique fixed point $\nabla^2_{\theta_1}Q_{\theta_1}$. Since
\begin{align}
    \nabla^2_{\theta_1}\bar{V}_{\theta_1}&(z',x') 
=
\sum\limits_{a'}\nabla_{\theta_1}\pi_{\theta_1}(a'|z',x')\nabla_{\theta_1}Q_{\theta_1}(z',x',a') \notag \\
&+ \sum\limits_{a'}\pi_{\theta_1}(a'|z',x')\nabla^2_{\theta_1}Q_{\theta_1}(z',x',a'),
\label{soft_value}
\end{align}
it follows that $\nabla^2_{\theta_1}\bar{V}_{\theta_1}$ exists.

From the fixed point characterization of $\nabla^2_{\theta_1}Q_{\theta_1}$ we obtain: 
\begin{align}
 &\left\Vert\nabla^2_{\theta_1}Q_{\theta_1}(z,x,a)\right\Vert 
      \leq \frac{1}{1-\beta}\Bigg(
     \left\Vert\nabla^2_{\theta_1}r_{\theta_1}(z,x,a) \right\Vert \notag\\
&+\beta \Big \Vert\sum\limits_{z'}\sigma_{\hat{\theta}_2}(z',z,x,a) \notag \\
&\times \sum\limits_{a'}\nabla_{\theta_1}\pi_{\theta_1}(a'|z',x')\nabla_{\theta_1}Q_{\theta_1}(z',x',a')\Big\Vert\Bigg). 
\label{bound_1}
\end{align}
Note that from the softmax structure of conditional choice probabilities,
\begin{align*}
         \nabla_{\theta_1}\pi_{\theta_1}(a|z,x)
     =&\pi_{\theta_1}(a|z,x)\Big[\nabla_{\theta_1}Q_{\theta_1}(z,x,a) \\
     &-\sum\limits_{a'}\pi_{\theta_1}(a'|z,x)\nabla_{\theta_1}Q_{\theta_1}(z,x,a')\Big].
\end{align*}
Hence, the second term on the right hand side of \eqref{bound_1} can be bounded by
\begin{align}
\Big \Vert \sum\limits_{z^{\prime }}\sigma _{\hat{\theta}_{2}}(z^{\prime
},z,x,a)&\sum\limits_{a^{\prime }}\nabla _{\theta _{1}}\pi _{\theta
_{1}}(a^{\prime }|z^{\prime },x^{\prime })\nabla _{\theta _{1}}Q_{\theta
_{1}}(z^{\prime },x^{\prime },a^{\prime })\Big \Vert  \notag\\
&\leq  \frac{2}{(1-\beta )^{2}}(L_{r,1})^{2}.
\label{bound_2}
\end{align}

Combining \eqref{bound_1} and \eqref{bound_2} we obtain:
\begin{align*}
 \left\Vert\nabla^2_{\theta_1}Q_{\theta_1}(z,x,a)\right\Vert 
      &\leq \frac{1}{1-\beta}L_{r,2}
+\frac{2\beta }{(1-\beta)^3}(L_{r,1})^2 = L_Q,
\end{align*}

Similarly, from \eqref{soft_value} we obtain:
\begin{align*}
&\left\Vert \nabla _{\theta _{1}}^{2}\bar{V}_{\theta _{1}}\right\Vert\\
&\leq \Big \Vert \sum\limits_{a^{\prime }}\nabla _{\theta _{1}}\pi _{\theta
_{1}}(a^{\prime }|z^{\prime },x^{\prime })\nabla _{\theta _{1}}Q_{\theta
_{1}}(z^{\prime },x^{\prime },a^{\prime }) \Big \Vert \\
&\hspace{10pt}+\Big \Vert
\sum\limits_{a^{\prime }}\pi _{\theta _{1}}(a^{\prime }|z^{\prime
},x^{\prime })\nabla _{\theta _{1}}^{2}Q_{\theta _{1}}(z^{\prime },x^{\prime
},a^{\prime }) \Big \Vert  \\
&\leq \Big\Vert \sum\limits_{a^{\prime }}\pi _{\theta _{1}}(a^{\prime
}|z^{\prime },x^{\prime })\big[\nabla _{\theta _{1}}Q_{\theta
_{1}}(z^{\prime },x^{\prime },a^{\prime }) -\sum\limits_{a^{\prime \prime}} \pi _{\theta _{1}}(a^{\prime \prime }|z^{\prime },x^{\prime }) \\
&\hspace{10pt}\times \nabla
_{\theta _{1}}Q_{\theta _{1}}(z^{\prime },x^{\prime },a^{\prime \prime })%
\big]\nabla _{\theta _{1}}Q_{\theta _{1}}(z^{\prime },x^{\prime },a^{\prime
}) \Big\Vert+L_{Q} \\
&\leq \frac{2}{(1-\beta )^{3}}(L_{r,1})^{2}+\frac{1}{1-\beta }L_{r,2} \\
&= L_{\bar{V}}. 
\end{align*}\endproof

\proof{Proof of Theorem \ref{convergenceThm}.}
Recall that
\begin{align*}
    &\nabla^2_{\theta_1}\hat{\ell}(\theta_1)=
     \sum\limits_{i=1}^N\sum\limits_{t=0}^{T-1} \nabla^2_{\theta_1} \log\pi_{\theta_1}(a_{t,i}|z_{t,i},x_{t,i})  \\
     & =
     \sum\limits_{i=1}^N\sum\limits_{t=0}^{T-1} \nabla^2_{\theta_1} Q_{\theta_1}(z_{t,i},x_{t,i},a_{t,i}) - \nabla^2_{\theta_1} \bar{V}_{\theta_1}(z_{t,i},x_{t,i}) 
\end{align*}

By Lemma \ref{bigLemma}, it follows that $\left\Vert\nabla^{2}_{\theta_1}\hat{\ell}(\theta_1)\right\Vert \leq L$ with
\[
L:=NT(L_Q+L_{\bar{V}})
\]
Or equivalently, $\nabla_{\theta_1}\hat{\ell}(\theta_1)$ is Lipschitz continuous in $\theta_1$ with constant $L$.

By Lipschitz continuous gradients, 
\begin{align*}
\hat{\ell}(\theta _{1}^{k+1})& \geq \hat{\ell}(\theta _{1}^{k})+\nabla \hat{%
\ell}(\theta _{1}^{k})^{\top }(\theta _{1}^{k+1}-\theta _{1}^{k})-\frac{L}{2}%
\left\Vert \theta _{1}^{k+1}-\theta _{1}^{k}\right\Vert ^{2} \\
& =\hat{\ell}(\theta _{1}^{k})+\rho \bigg(1-\frac{\rho L }{2}\bigg)\left\Vert \nabla
_{\theta _{1}}\hat{\ell}(\theta _{1}^{k})\right\Vert ^{2}.
\end{align*}

Hence,%
\begin{equation*}
\rho \bigg(1-\frac{\rho L }{2}\bigg)\left\Vert \nabla _{\theta _{1}}\hat{\ell}(\theta
_{1}^{k})\right\Vert ^{2}\leq \hat{\ell}(\theta _{1}^{k+1})-\hat{\ell}%
(\theta _{1}^{k}).
\end{equation*}%
Adding over $k=1,\ldots ,K$ we obtain%
\begin{equation*}
\rho \bigg(1-\frac{\rho L }{2}\bigg)\sum\limits_{k=1}^{K}\left\Vert \nabla _{\theta
_{1}}\hat{\ell}(\theta _{1}^{k})\right\Vert ^{2}\leq \hat{\ell}(\theta
_{1}^{K})-\hat{\ell}(\theta _{1}^{0})\leq \hat{\ell}(\theta _{1}^{\ast })-%
\hat{\ell}(\theta _{1}^{0})
\end{equation*}%
where $\theta _{1}^{\ast }$ is a maximizerof log-likelihood. It follows that%
\begin{align*}
\min_{k\in \{1,\ldots ,K\}}\left\Vert \nabla _{\theta _{1}}\hat{\ell}(\theta
_{1}^{k})\right\Vert ^{2}&\leq \frac{1}{K}\sum\limits_{k=1}^{K}\left\Vert
\nabla _{\theta _{1}}\hat{\ell}(\theta _{1}^{k})\right\Vert ^{2}\\
&\leq \frac{1%
}{\rho \Big(1-\frac{\rho L }{2}\Big)}\frac{\hat{\ell}(\theta _{1}^{\ast })-\hat{\ell}%
(\theta _{1}^{0})}{K}.
\end{align*}
\endproof

\subsection{Identification Results}

\proof{Proof of Theorem \ref{DynamicsIdentification}.}
For any given two system dynamics $P_{\theta_2}(z',z,a)$, $P_{\theta_2^{'}}(z',z,a)$, where $P(z',z,a)=\{P_{ij}(z',z,a)\}, P_{ij}(z',z,a)= P(z_{t+1}=z', s_{t+1}=j | z_t=z, s_t = i, a_t=a)$,  we show that they can be distinguished by the data. Since the dataset contains $x_0, z_0, a_0$ and $|S|$ is known, we can obtain
\begin{align*}
    \sigma_{\theta_2}^0(z_1,z_0, x_0, a_0) &= \sum_{s_{1}}\sum_{s}x_0(s)P_{\theta_2}(z_1,s_1|z_0,s,a_0)\\ &=\sum_{s}x_0(s)P_{\theta_2}(z_1|z_0,s,a_0),
\end{align*}
and 
\begin{align*}
    \sigma_{\theta_2^{'}}^0(z_1, z_0, x_0, a_0) &=\sum_{s_{1}}\sum_{s}x_0(s)P_{\theta_2^{'}}(z_1,s_1|z_0,s,a_0) \\
    &=\sum_{s}x_0(s)P_{\theta_2^{'}}(z_1|z_0, s,a_0).
\end{align*}
Note that $$\sigma^0(z_1,z_0,x_0,a_0) = \hat{P}(z_1|\zeta_0,a_0),$$ where $\hat{P}(z_1|\zeta_0,a_0)$ is a function of the first period data. Thus, $\sigma$ can be obtained from the data and $\sigma_{\theta_2}^0=\sigma_{\theta_2^{'}}^0$ if and only if $P_{\theta_2}(z'|z,s,a)=P_{\theta_2^{'}}(z'|z,s,a)$. If $\sigma_{\theta_2}^0 \neq \sigma_{\theta_2^{'}}^0$, we are done. However, it is possible that there exists $s'$ such that $P_{\theta_2}(z',s'|z, s,a)\neq P_{\theta_2^{'}}(z',s'|z, s,a)$ but $P_{\theta_2}(z'|z,s,a)=P_{\theta_2^{'}}(z'|z,s,a)$. In this case, update belief by Eq. \eqref{lambda}, $$x_{1, \theta_2} = \lambda_{\theta_2}(z_1,z_0,x_0,a_0) = \frac{x_0P_{\theta_2}(z_1,z_0,a_0)}{\sigma_{\theta_2}^0(z_1,z_0, x_0, a_0)},$$ $$x_{1,\theta_2^{'}} = \lambda_{\theta_2^{'}}(z_1,z_0,x_0,a_0) = \frac{x_0P_{\theta_2^{'}}(z_1,z_0,a_0)}{\sigma_{\theta_2^{'}}^0(z_1,z_0, x_0, a_0)}.$$ Then $x_{1,\theta_2} = x_{1,\theta_2^{'}}$ if and only if $P_{\theta_2}(z',z,a) = P_{\theta_2^{'}}(z',z,a)$. Now, $\sigma^1_{\theta_2}(z_2, z_1, x_{1,\theta_2}, a_1)=\sum_{s}x_{1,\theta_2}(s)P_{\theta_2}(z_2|z_1,s,a_1)$ and $\sigma^1_{\theta_2^{'}}(z_2, z_1, x_{1,\theta_2^{'}}, a_1)=\sum_{s}x_{1,\theta_2^{'}}(s)P_{\theta_2^{'}}(z_2|z_1,s,a_1)$, and again $\sigma^1$ is obtainable from the two periods of data as $$\sigma^1(z_2,z_1,x_1,a_1) = \hat{P}(z_2|\zeta_1,a_1).$$ Now, $\sigma^1_{\theta_2}=\sigma^1_{\theta_2^{'}}$ if and only if $x_{1,\theta_2}=x_{1,\theta_2^{'}}$, indicating $P_{\theta_2}(z',z,a) = P_{\theta_2^{'}}(z',z,a)$ assuming $P(z',z,a)$ is not rank-1. 
\endproof

\proof{Proof of Theorem \ref{Identification}.}
By Theorem  \ref{SufficientTheorem} and Theorem \ref{DynamicsIdentification}, both $\pi(a|z_t, x_t)=\hat{P}(a|\zeta_t)$ and hidden dynamics $\{P_{\theta_2}(z',z,a)\}$ can be identified from the data. Treating belief as the state and since $|\zeta_t| < \infty$, Proposition 1 in \cite{Hotz} and \cite{MagnacThesmar2002} show that there is a one-to-one mapping $q(Z,X): R^{|A|}\rightarrow R^{|A|}$, only depending on $\mu$, which maps the choice probability set $\{\pi_{\theta}(a|z,x)\}$  to the set of the difference in action-specific value function $\{Q_{\theta}(z,x,a)-Q_{\theta}(z,x,a^0)\}$, namely, 
\begin{align}
  Q_{\theta}(z,x,a)-Q_{\theta}(z,x,a^0)=q_a(\{\pi_{\theta}(a'|z,x)\}; \mu),  \label{Qdiff}
\end{align}
$q_0(.)=0$, and $q=(q_0,...,q_{|A|-1})$. Thus, if we know $Q_{\theta}(z,x,a^0)$, we can recover $Q_{\theta}(z,x,a),  \forall a \in A \backslash a^0$. 
Note that 
\begin{alignat*}{2}
    &Q_{\theta}(z,x,a) = r_{\theta_1}(z,x,a) +\beta E_{z'|z,x,a}\Big[ E_{\epsilon'|z'} \max_{a' \in A} \big\{ ...\notag\\
          \end{alignat*}
  \begin{alignat}{2}
    &\hspace{10pt} Q_{\theta}(z',\lambda_{\theta_2}(z',z,x,a), a') + \epsilon'(a')\big \} \Big ] \notag\\
    &= r_{\theta_1}(z,x,a)  \notag\\
    &\hspace{10pt}+ \beta E_{z'|z,x,a}\Bigg[ E_{\epsilon'|z'} \Big[\max_{a' \in A}\big \{Q_{\theta}(z',\lambda_{\theta_2}(z',z,x,a), a') \notag \\
    &\hspace{60pt}-Q_{\theta}(z',\lambda_{\theta_2}(z',z,x,a), a^0) + \epsilon'(a')\big \}\notag\\
    &\hspace{60pt}+Q_{\theta}(z',\lambda_{\theta_2}(z',z,x,a), a^0)\Big ] \Bigg ] \\
     &= r_{\theta_1}(z,x,a) \notag \\
     &\hspace{10pt} + \beta E_{z'|z,x,a}\Big[E_{\epsilon'|z'} \max_{a' \in A}\big \{Q_{\theta}(z',\lambda_{\theta_2}(z',z,x,a), a') \notag\\
    &\hspace{60pt}- Q_{\theta}(z',\lambda_{\theta_2}(z',z,x,a), a^0) + \epsilon'(a')\big \} \Big] \notag\\
     &\hspace{10pt}+ \beta E_{z'|z,x,a}\left [Q_{\theta}(z',\lambda_{\theta_2}(z',z,x,a), a^0) \right] 
     \label{Iden}
\end{alignat}

Because of the mapping $q$ in Eq. \eqref{Qdiff}, $\pi_{\theta}(a|z_t, x_t)=\hat{P}(a|\zeta_t)$, and that both the hidden dynamic $\{P_{\theta_2}(z',z,a)\}$ and $\mu$ are known, the quantity 
\begin{align*}
 C=\beta E_{z'|z,x,a}\Big[&E_{\epsilon'|z'} \max_{a' \in A}\big \{Q_{\theta}(z',\lambda_{\theta_2}(z',z,x,a), a')\notag\\ &-Q_{\theta}(z',\lambda_{\theta_2}(z',z,x,a), a^0) + \epsilon'(a')\big \} \Big]   
\end{align*}
 is known. Under the assumption of $r_{\theta_1}(Z,S,a^0)=0$, we have
\begin{align}
    Q_{\theta}(z,x,a^0) = C+ \beta E_{z'|z,x,a^0} [Q_{\theta}(z',\lambda_{\theta_2}(z',z,x,a^0), a^0)]\label{Qx00}
\end{align}
with only unknown $Q_{\theta}(Z,X,a^0)$. It is easy to see there is a unique solution to Eq. \eqref{Qx00} due to the contraction mapping theorem. Consequently, all $Q_{\theta}(Z,X,a), a \in A$ can be recovered by Eq. \eqref{Qdiff}. Lastly,   
\begin{align*}
 r_{\theta_1}(z,x,a) =& Q_{\theta}(z,x,a) -C\notag\\
 &-\beta E_{z'|z,x,a^0}\left [Q_{\theta}(z',\lambda_{\theta_2}(z',z,x,a^0), a^0) \right] . 
\end{align*}
As $$r_{\theta_1}(z,x,a)=\sum_{s \in S}x(s)r_{\theta_1}(z,s,a), \{r_{\theta_1}(s,a): s \in S, a \in A\}$$ can be uniquely determined. 
\endproof
\proof{Proof of Corollary \ref{corollary1}.}
When $Q_{T, \theta}(Z,X,a^0)$ and $r_{\theta_1}(Z,S,a^0)$ are known, we can obtain $Q_{t,\theta}(Z,X,a^0)$ via
\begin{align}
    &Q_{t,\theta}(z,x,a^0) =r_{\theta_1}(z,x,a^0) \notag \\
    &+\beta E_{z'|z,x,a^0}\Big[E_{\epsilon'|z'} \max_{a' \in A}\big \{Q_{t+1,\theta }(z',\lambda_{\theta_2}(z',z,x,a^0), a') \notag\\
    &\hspace{60pt}+ \epsilon'(a') - Q_{t+1,\theta}(z',\lambda_{\theta_2}(z',z,x,a^0), a^0) \big \} \Big]\notag\\
     & + \beta E_{z'|z,x,a^0}\left [Q_{t+1, \theta}(z',\lambda_{\theta_2}(z',z,x,a^0), a^0) \right] 
\end{align}
The rest follows exactly as in the proof of Theorem \ref{Identification}.
\endproof

\proof{Proof of Theorem \ref{x0unknown}.} 
The proof of Theorem \ref{DynamicsIdentification} shows that $\theta_2$ can be uniquely determined by 
$$\sigma_{\theta_2}(z_{t+1},z_t,\lambda_{\theta_2}(z_t,z_{t-1},x_{t-1}^*,a_{t-1}),a_t) = \hat{P}(z_{t+1}|\zeta_t, a_t),$$
given the true value $x_{t-1}^*$ is known for  $\zeta_t$. Namely, there exists a unique $\theta_2^*$ such that
\begin{align*}
    \sigma_{\theta_2^*}(z_{t+1},z_t,\lambda_{\theta_2^*}(z_t,z_{t-1},x_{t-1}^*,a_{t-1}),a_t) =  \hat{P}(z_{t+1}|\zeta_t, a_t),
\end{align*}
and $\forall \theta_2 \neq \theta_2^*$, there is an $\epsilon>0$ such that 
\begin{align*}
    D(\sigma_{\theta_2}(.,z_t,\lambda^M_{\theta_2}(z_{t-M}^{t}, a_{t-M}^{t-1}, x_{t-M}^*),a_t),  \hat{P}(.|\zeta_t,a_t)) \geq \epsilon,
\end{align*}
where $x_{t-M}^*$ is the true belief at $t-M$. Due to the contraction coefficient $\eta(P_{\theta_2}(z',z,a))<1$, we have
$$D(\lambda^M_{\theta_2}(z_{t-M}^{t}, a_{t-M}^{t-1}, x_{t-M}^*), \lambda^M_{\theta_2}(z_{t-M}^{t}, a_{t-M}^{t-1}, x_{t-M})) \leq \eta^M,$$ 
where $\eta = \max_{z',z',a} \eta(P_{\theta_2}(z',z,a)) <1$ since $|Z|<\infty, |A| <\infty$. Thus, $\forall \theta_2 \neq \theta_2^*$, we have 
\begin{align*}
&\lim_{M \rightarrow \infty}   D(\sigma_{\theta_2}(.,z_t,\lambda^M_{\theta_2}(z_{t-M}^{t}, a_{t-M}^{t-1}, x_{t-M}), a_t), \hat{P}(.|\zeta_t, a_t)) \\
&=\lim_{M \rightarrow \infty}   D(\sigma_{\theta_2}(.,z_t, \lambda^M_{\theta_2}(z_{t-M}^{t}, a_{t-M}^{t-1}, x_{t-M}^*), a_t), \hat{P}(.|\zeta_t, a_t)) \\
&\geq \epsilon, \forall  x_{t-M} \in X,
\end{align*}
indicating $\theta_2$ can be distinguished by the data given $M$ is sufficiently large. 
Once $\theta_2$ is determined, $\theta_1$ can be determined by Theorem \ref{Identification}, which completes the proof. 
\endproof


\bibliographystyle{ieeetr}
\bibliography{PaperBIB}
\vskip 0pt plus -1fil

\begin{IEEEbiography}
	[{\includegraphics[width=1.05in,height=1in,clip,keepaspectratio]{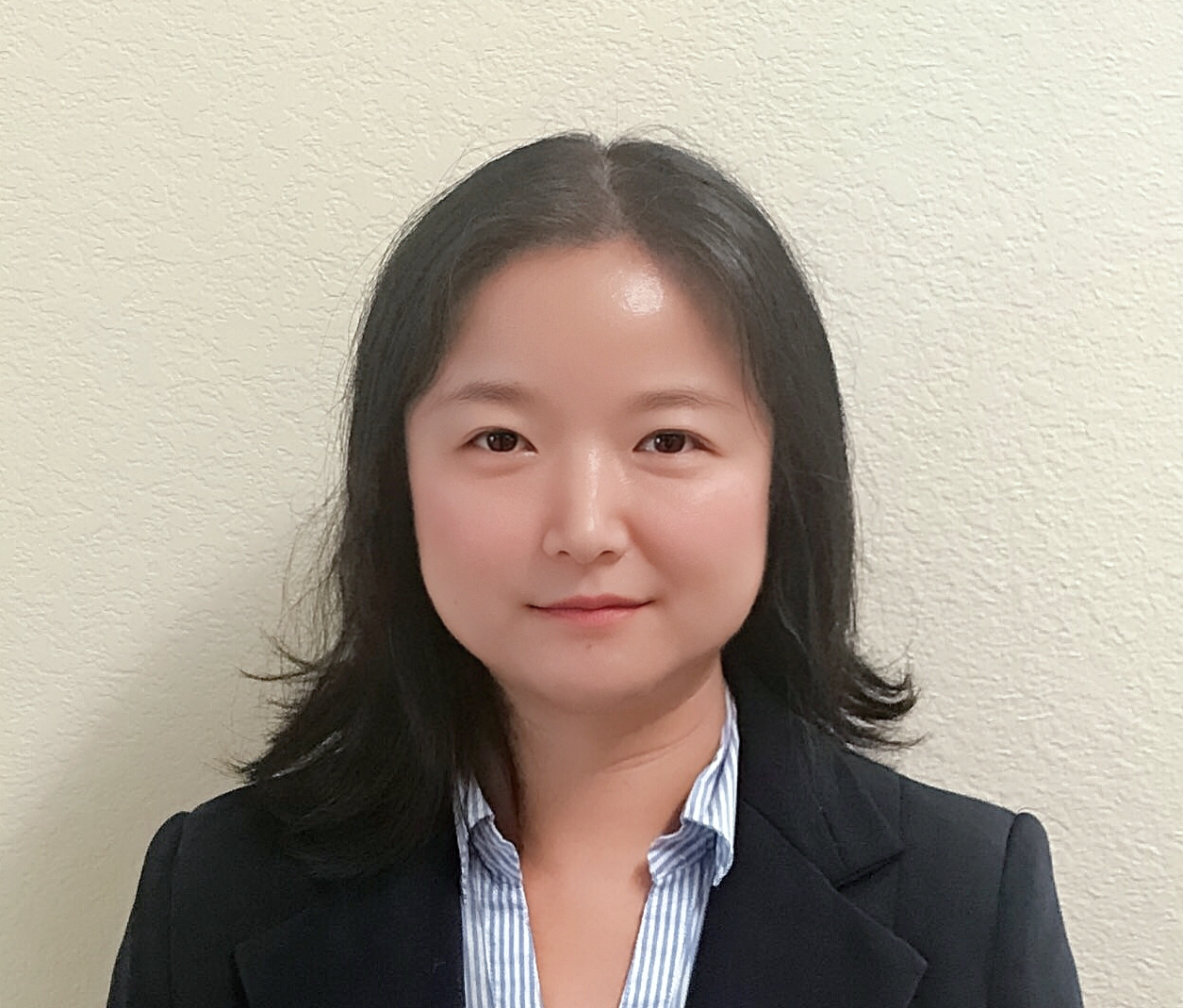}}] 
	{Yanling Chang} is an assistant professor in the Department of Engineering Technology and Industrial Distribution and the Department of Industrial and Systems Engineering, Texas A\&M University, College Station, TX, USA. She received her Bachelor’s degree in Electronic and Information Science and Technology from Peking University, Beijing, China; the Master's degree in Mathematics and the Ph.D. degree in Operations Research both from Georgia Institute of Technology, Atlanta, GA, USA. Her research interests include partially observable Markov decision processes, game theory, and optimization.  
\end{IEEEbiography}
\vskip 0pt plus -1fil

\begin{IEEEbiography}
[{\includegraphics[width=1.05in,height=1.25in,clip,keepaspectratio]{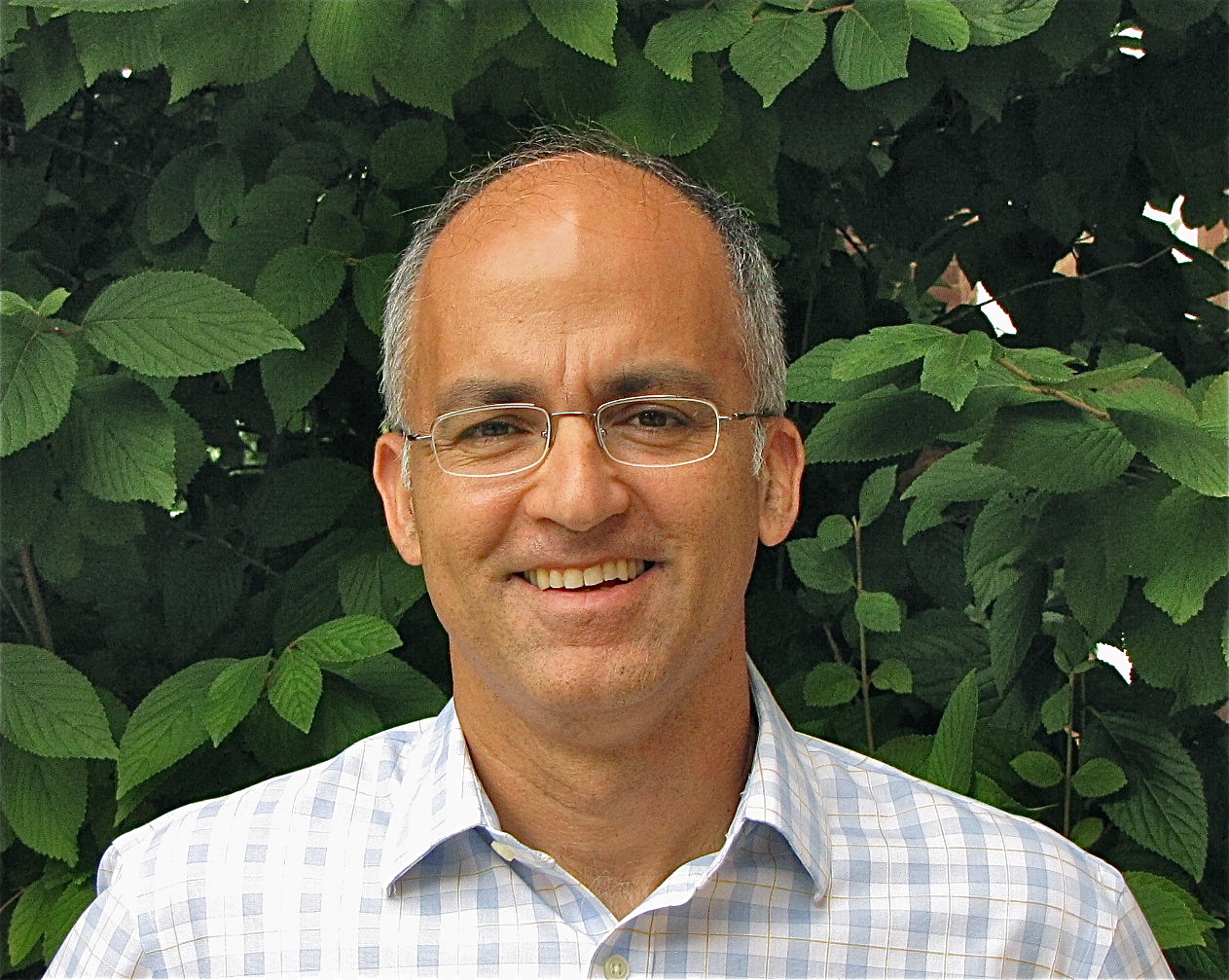}}] 
{Alfredo Garcia} received the Degree in electrical engineering from the Universidad de los Andes, Bogot\'a,Colombia,in 1991,the Dipl\^ome d’Etudes Approfondies in automatic control from the Universit\'e Paul Sabatier, Toulouse, France, in 1992, and the Ph.D. degree in industrial and operations engineering from the University of Michigan, Ann Arbor, MI, USA, in 1997.
From 1997 to 2001, he was a consultant to government agencies and private utilities in the electric power industry. From 2001 to 2015, he
was a Faculty with the Department of Systems and Information Engineering, University of Virginia, Charlottesville, VA, USA. From 2015 to 2017, he was a Professor with the Department of Industrial and Systems Engineering, University of Florida, Gainesville, FL, USA. In 2018, he joined the Department of Industrial and System Engineering, Texas A\&M University, College Station, TX, USA. His research interests include game theory and dynamic optimization with applications in communications and energy networks.
\end{IEEEbiography}
\vskip 0pt plus -1fil

\begin{IEEEbiography}
	[{\includegraphics[width=1.05in,height=1in,clip,keepaspectratio]{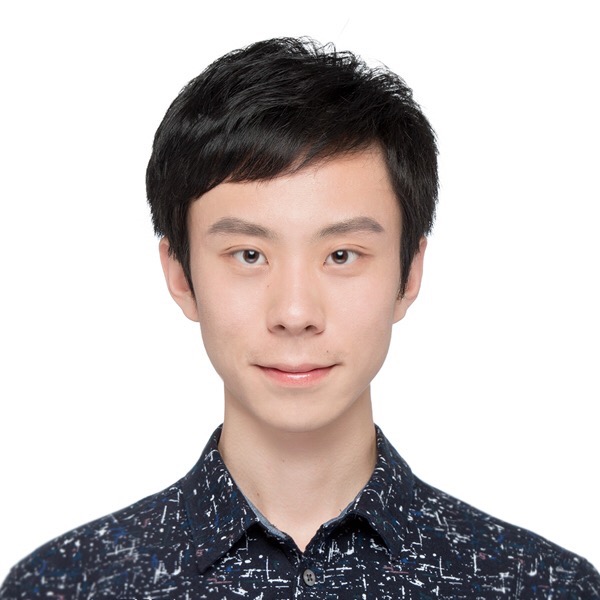}}] 
	{Zhide Wang} is a PhD candidate in the Department of Industrial and System Engineering, Texas A\&M University, College Station, TX, USA. He received his Bachelor's degree in Industrial Engineering and Management from Shanghai Jiao Tong University, Shanghai, China. His research interests include partially observable Markov decision processes and structural estimation of dynamic discrete choice models with applications in cognitive psychology.
\end{IEEEbiography}
\vskip 0pt plus -1fil

\begin{IEEEbiography}
	[{\includegraphics[width=1.05in,height=1in,clip,keepaspectratio]{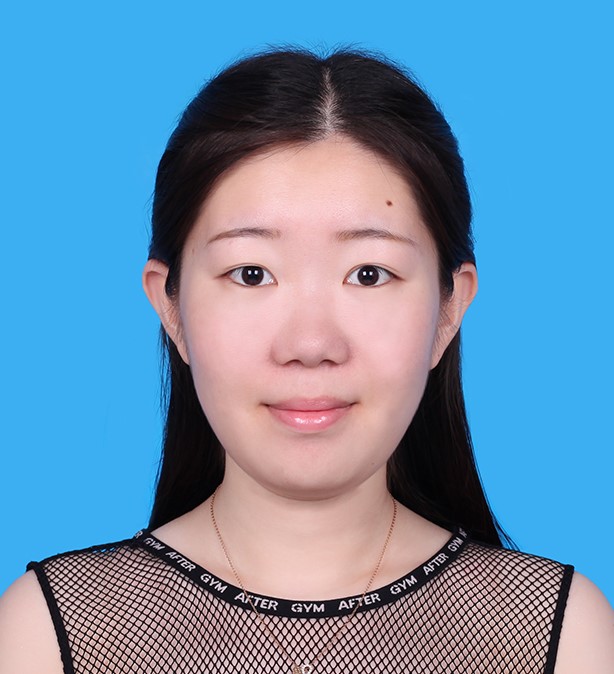}}] 
	{Lu Sun} is a PhD student in the Department of Industrial and System Engineering, Texas A\&M University, College Station, TX, USA. She received her Bachelor's degree in Mathematics from Beihang University, Beijing, China. Her research interests include partially observable Markov decision processes and game theory.
\end{IEEEbiography}
\end{document}